\documentclass[conference]{IEEEtran}
\usepackage{cite}
\usepackage{amsmath}
\usepackage{mathtools}
\usepackage{amssymb}
\usepackage{amsfonts}

\usepackage{amsthm}
\usepackage[linesnumbered]{algorithm2e}
\usepackage{graphicx}
\usepackage{textcomp}
\usepackage{enumitem}
\usepackage{url}
\usepackage{floatrow}
\usepackage{subcaption}
\ifCLASSINFOpdf
\else
\fi
\begin{document}
%
\title{\textsf{nn-dependability-kit}: Engineering Neural Networks for Safety-Critical Autonomous Driving Systems}


%
\author{\IEEEauthorblockN{Chih-Hong Cheng,
Chung-Hao Huang and
Georg N\"{u}hrenberg} 
\IEEEauthorblockA{fortiss - Research Institute of the Free State of Bavaria\\
Munich, Germany \\  
\texttt{https://github.com/dependable-ai/nn-dependability-kit}}
}

\IEEEspecialpapernotice{(Invited Paper)}

\maketitle

\begin{abstract}

Can engineering neural networks be approached in a disciplined way similar to how engineers build software for civil aircraft? We present \textsf{nn-dependability-kit}, an open-source toolbox to support safety engineering of neural networks for autonomous driving systems. The rationale behind \textsf{nn-dependability-kit} is to consider a structured approach (via Goal Structuring Notation) to argue the quality of neural networks. In particular, the tool realizes recent scientific results including (a) novel dependability metrics for indicating sufficient elimination of uncertainties in the product life cycle, (b) formal reasoning engine for ensuring that the generalization does not lead to undesired behaviors, and (c) runtime monitoring for reasoning whether a decision of a neural network in operation is supported by prior similarities in the training data. A proprietary version of \textsf{nn-dependability-kit} has been used  to improve the quality of a level-3 autonomous driving component developed by Audi for highway maneuvers. 
\end{abstract}


%
\IEEEpeerreviewmaketitle

\section{Introduction}

In recent years, neural networks have been widely adopted in engineering automated driving systems with examples in  perception, decision making, or even end-to-end scenarios. As these systems are safety-critical in nature, problems during operation such as failed identification of pedestrians may contribute to risky behaviors. Importantly, the root cause of these undesired behaviors can be independent of hardware faults or software programming errors but \emph{can solely reside in the data-driven engineering process}, e.g., unexpected results of function extrapolation between correctly classified training data. 

In this paper, we present \textsf{nn-dependability-kit}, an open-source toolbox to support data-driven engineering of neural networks for safety-critical domains. The goal is to provide evidence of uncertainty reduction in key phases of the product life cycle, ranging from data collection, training \& validation, testing \& generalization, to operation. \textsf{nn-dependability-kit} is built upon our  previous research work\cite{cheng2017maximum,cheng2018towards,DBLP:conf/atva/ChengHY18,DBLP:journals/corr/abs-1809-06573,DBLP:conf/date/ChengDHHN0RT18}, where (a) novel dependability metrics~\cite{cheng2018towards,DBLP:conf/atva/ChengHY18} are introduced to indicate  uncertainties being reduced in the engineering life cycle, (b) formal reasoning engine~\cite{cheng2017maximum,DBLP:conf/date/ChengDHHN0RT18} is used to ensure that the generalization does not lead to undesired behaviors, and (c) runtime neuron activation pattern monitoring~\cite{DBLP:journals/corr/abs-1809-06573} is applied to reason whether a decision of a neural network in operation time is supported by prior similarities in the training data.


\begin{figure*}[t]
	\includegraphics[width=0.8\columnwidth, trim=4.5cm 1cm 4.5cm 0.5cm, clip]{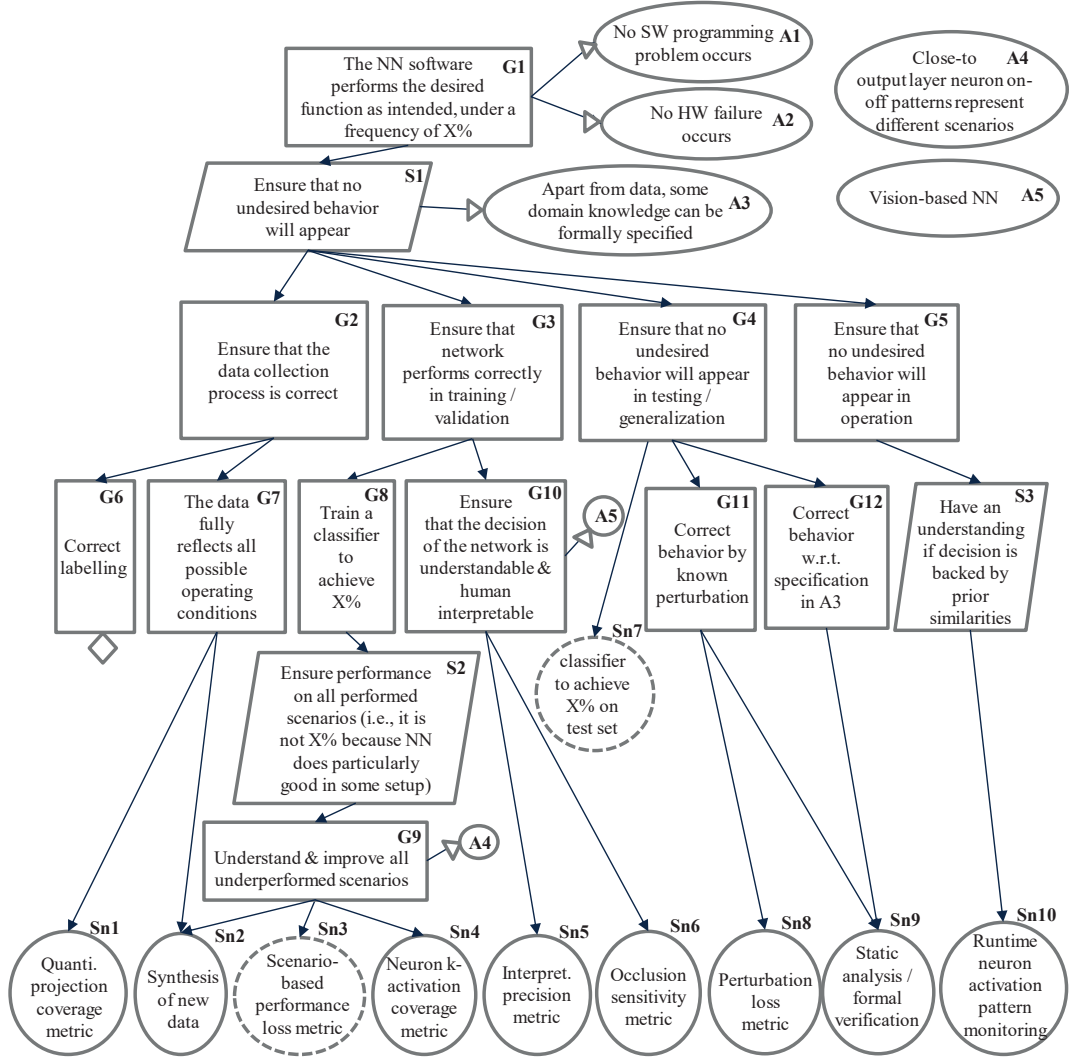}
	\caption{Using a simplified GSN to understand solutions~(Sn) provided by \textsf{nn-dependability-kit}, including the associated goals~(G), assumptions~(A) and strategies~(S).}%
	\label{fig:gsn}
\end{figure*}

Concerning related work, our results~\cite{cheng2017maximum,cheng2018towards,DBLP:conf/atva/ChengHY18,DBLP:journals/corr/abs-1809-06573,DBLP:conf/date/ChengDHHN0RT18} are within recent research efforts from the software engineering and formal methods community targeting to provide provable guarantee over neural networks~\cite{weng2018towards,DBLP:conf/cav/KatzBDJK17,pulina2010abstraction,cheng2017maximum,kolter2017provable,ehlers2017formal,DBLP:conf/sp/GehrMDTCV18,dvijotham2018dual,dutta2018output, DBLP:journals/corr/SeshiaS16,wang2018formal} or to test neural networks~\cite{sun2018testing,sun2018concolic,DBLP:conf/atva/ChengHY18,pei2017deepxplore,DBLP:journals/corr/abs-1809-09310,DBLP:conf/kbse/MaJZSXLCSLLZW18,pezzementi2018putting,hutchison2018robustness}. The static analysis engine inside \textsf{nn-dependability-kit}  for formally analyzing neural networks, as introduced in our work in 2017~\cite{cheng2017maximum} as a pre-processing step before exact constraint solving, has been further extended to support the octagon abstract domain~\cite{mine2006octagon} in addition to using the interval domain. One deficiency of the above-mentioned works is how to connect safety goals or uncertainty identification~\cite{DBLP:conf/safecomp/GauerhofMB18,DBLP:conf/safecomp/KlasV18,DBLP:conf/safecomp/ShafaeiKOK18} to concrete evidences required in the safety engineering process. This gap is tightened by our earlier work of dependability metrics~\cite{cheng2018towards} being partly integrated inside \textsf{nn-dependability-kit}. Lastly, our runtime monitoring technique~\cite{DBLP:journals/corr/abs-1809-06573} is different from known results that either use additional semantic embedding (and integrate it in the loss function) for computing difference measures~\cite{mandelbaum2017distance} or use Monte-Carlo dropout~\cite{gal2016dropout} as ensembles: our approach provides a sound guarantee of the similarity measure based on the neuron word distance to the training data.

In terms of practicality, a proprietary version of \textsf{nn-dependability-kit}  has been successfully used  to understand the quality of a level-3 autonomous highway maneuvering component developed by Audi.

\section{Features and Connections to Safety Cases}

  Figure~\ref{fig:gsn} provides a simplified Goal Structuring Notation (GSN)~\cite{kelly2004goal} diagram to assist in understanding how features provided by \textsf{nn-dependability-kit}  contribute to the overall safety goal\footnote{Note that the GSN in Figure~\ref{fig:gsn} may not be complete, but it can serve as a basis for further extensions.}. Our proposed metrics, unless explicitly specified, are based on extensions of our early work~\cite{cheng2018towards}. Starting with the goal of having a neural network to function correctly~(G1), based on assumptions where no software and hardware fault appears (A1, A2), the strategy~(S1) is to ensure that within different phases of the product life cycle, correctness is ensured. These phases include data preparation~(G2), training and validation~(G3), testing and generalization~(G4), and operation~(G5).

\begin{description}
	\item[(Data preparation)] Data preparation includes \emph{data collection} and \emph{labeling}. Apart from correctly labeling the data~(G6), one needs to ensure that the collected data covers all operating scenarios~(G7). An artificial example of violating such a principle is to only collect data in sunny weather, while the vehicle is also expected to be operated in snowy weather. Quantitative projection coverage metric~\cite{DBLP:conf/atva/ChengHY18} and its associated test case generation techniques (Sn1, Sn2), based on the concept of combinatorial testing~\cite{lawrence2011survey,colbourn2004combinatorial}, are used to provide a relative form of completeness against the combinatorial explosion of scenarios. 

\item[(Training and validation)] 
Currently, the goal of training correctness is refined into two subgoals of understanding the decision of the neural network~(G10) and correctness rate~(G8), which is further refined by considering the performance under different operating scenarios~(G9).
Under the assumption where the neural network is used for vision-based object detection~(A5), metrics such as interpretation precision~(Sn5) and sensitivity of occlusion~(Sn6) are provided by \textsf{nn-dependability-kit}.

\item[(Testing and generalization)] Apart from classical performance measures (Sn7), we also test the generalization subject to known perturbations such as haze or Gaussian noise~(G11) using the \emph{perturbation loss metric} (Sn8). Provided that domain knowledge can be formally specified~(A3), one can also apply formal verification~(Sn9) to examine if the neural network demonstrates correct behavior with respect to the specification~(G12).  

\begin{figure*}[t]
    \centering
    \includegraphics[width=0.95\textwidth]{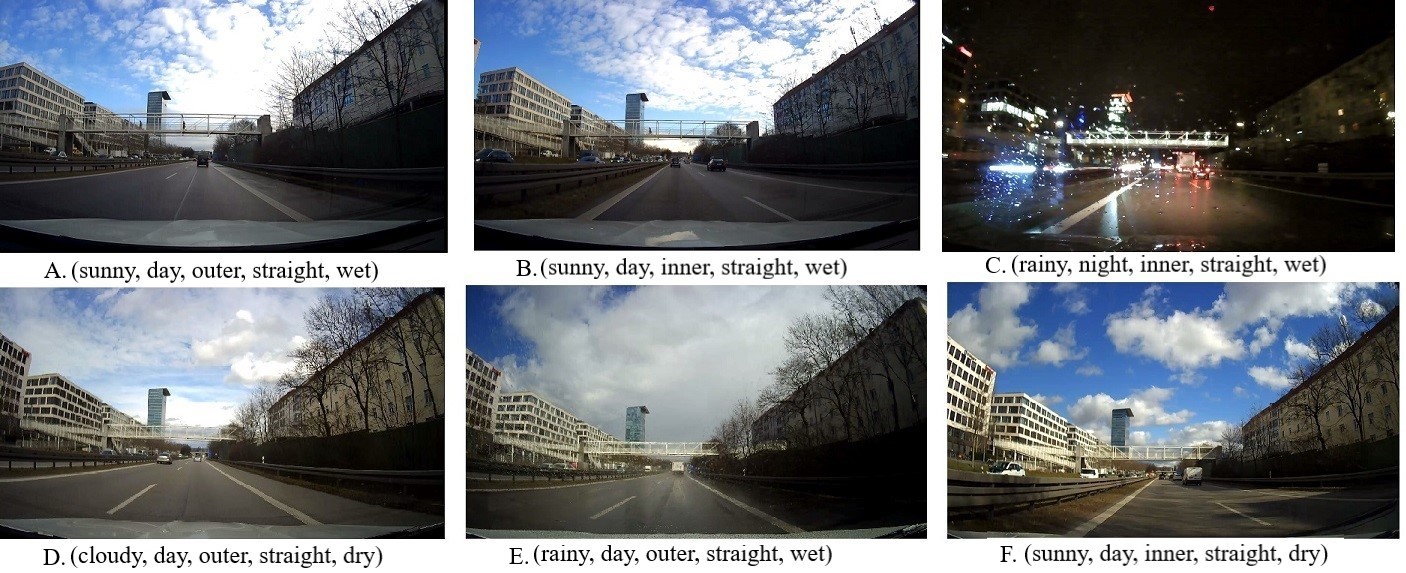}
    \caption{Photos of the same location (near Munich Schwabing) with variations in terms of  lighting conditions (day, night),   road surface situations (dry, wet),  currently located lanes (inner, outer), and   weather conditions (sunny, cloudy, rainy).}
    \label{fig:A9}
\end{figure*} 
\begin{figure*}[t]
    \centering
    \includegraphics[width=\textwidth]{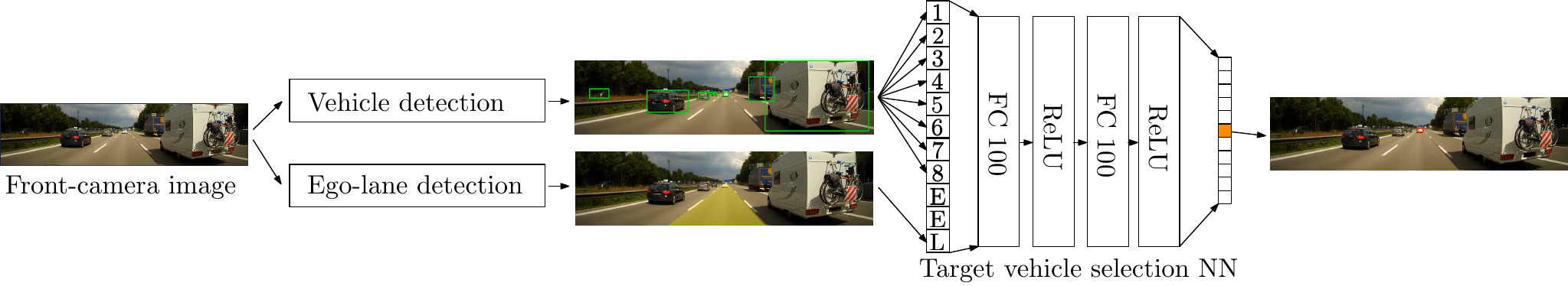}
    \caption{Illustration of the \emph{target vehicle selection} pipeline for ACC. The input features of the Target vehicle selection NN are defined as follows: 1-8 (possibly up to~$10$) are bounding boxes of detected vehicles, E is an empty input slot, i.e., there are less than ten vehicles, and L stands for the ego-lane information.}
    \label{fig:acc}
\end{figure*}

\item[(Operation)] In operation time, as the ground truth is no longer available, a dependable system  shall raise warnings when a decision of the neural network is not supported by prior similarities in training (S3).   \textsf{nn-dependability-kit} provides runtime monitoring (Sn10)~\cite{DBLP:journals/corr/abs-1809-06573} by first using binary decision diagrams (BDD)~\cite{bryant1992symbolic} to record binarized neuron activation patterns in training time, followed by checking if an activation pattern produced during operation is contained in the BDD. 
\end{description}

\section{Using \textsf{nn-dependability-kit}} 
 
In this section, we highlight the usage of  \textsf{nn-dependability-kit} with three examples; interested readers can download these examples as jupyter notebooks from the tool website. 
 
\paragraph{Systematic data collection for training and testing}  The first example is related to the completeness of the data used in testing autonomous driving components. The background is that one wishes to establish a coverage-driven data collection process where data diversity is approached in a disciplined manner. This is in contrast to unsystematic approaches such as merely driving as many kilometers as possible; the visited scenarios under unsystematic approaches may be highly repetitive.

Here for simplicity, we only consider five \emph{discrete categories} to evaluate the diversity of the collected data. 

\begin{description}
   \item[weather] $C_1 = \{\textrm{cloudy}, \textrm{rainy}, \textrm{sunny}\}$
  \item[day] $C_2 = \{\textrm{day}, \textrm{night}\}$
  \item[vehicle current lane] $C_3 = \{\textrm{inner}, \textrm{outer}\}$
   \item[curvature] $C_4 = \{\textrm{straight},  \textrm{left\_bending},  \textrm{right\_bending}\}$ 
     \item[surface type] $C_5 = \{\textrm{dry}, \textrm{wet}\}$
\end{description}

With discrete categorization, images or videos sharing the same semantic attributes fall into the same equivalence class. It is immediate that the number of equivalence classes is exponential to number of discrete categories (in this example, the number of equivalence classes equals $3\times2\times2 \times 3 \times 2 = 72$). Figure~\ref{fig:A9} provides~$6$ images taken from the German A9 highway (near Munich Schwabing); each of them is an instance of a particular equivalence class. With~$6$ images, the coverage by considering covering every equivalence class with one image or video is only $\frac{6}{72} = 8.3\%$. For pragmatic situations, the number of discrete categories can easily reach~$40$, making the minimum number of test cases to achieve full coverage be thousands of billions~($2^{40}$). 


\begin{figure*}[t]
	\includegraphics[width=0.75\columnwidth, trim=4.5cm 2cm 4cm 2cm, clip]{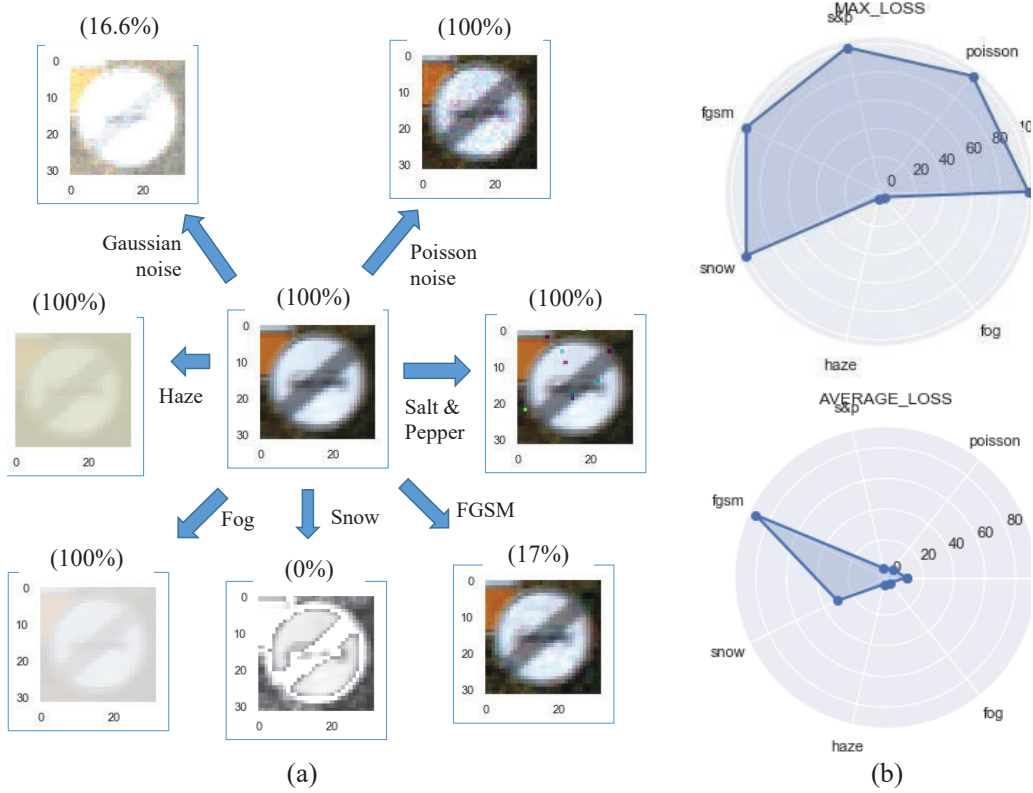}
	\caption{Using \textsf{nn-dependability-kit} to perturb the image of a traffic sign (a) and compute the maximum and average performance drop, by applying perturbation on the data set (b).}%
	\label{fig:perturbation}
\end{figure*}

In \textsf{nn-dependability-kit}, the \emph{scenario k-projection coverage} tries to create a\emph{ rational  yet relatively complete} coverage criteria that avoids the above mentioned combinatorial explosion (rational - possible to achieve $100\%$) while still guaranteeing diversity of test data (relatively complete). For example, when setting $k=2$, the $2$-projection coverage claims~$100\%$ coverage so long as the collected data is able to cover arbitrary \emph{pairs} of conditions from two different categories. In this example, the collected data should cover every pair $(c_i, c_j)$ where $i, j \in \{1,2,3,4,5\}$, $i\neq j$, $c_i \in C_i$ and $c_j \in C_j$. This makes the number of test cases to achieve full coverage be quadratic (when $k=2$)  to the number of categories. 

\textsf{nn-dependability-kit} reports that a data set using~$6$ images in Figure~\ref{fig:A9} achieves a $2$-projection coverage of $\frac{32}{57}$. In addition, the solver automatically suggests to include $(\textrm{cloudy}, \textrm{night}, \textrm{inner}, \textrm{right\_bending}, \textrm{dry})$ as the next image or video to be collected; the suggested case  can maximally increase the coverage by $\frac{7}{57}$. The denominator~$57$ comes from the following computation $\sum_{i < j, i,j \in \{1,2,3,4,5\}} |C_i|\times|C_j|$. One may additionally include constraints to describe  unreasonable cases that cannot be found in reality. For example, constraints such as $0 \leq \textrm{C1.sunny} + \textrm{C2.night} \leq 1$ prohibits the solver to consider proposals where the image is taken
at night ($\textrm{C2.night} = 1$) but the weather is sunny ($\textrm{C1.sunny} = 1$).  



\paragraph{Formal verification of a highway front car selection network}  
The second example is to formally verify properties of a neural network that selects the \textit{target vehicle} for an adaptive cruise control (ACC) system to follow. The overall pipeline is illustrated in Figure~\ref{fig:acc}, where two modules use images of a front facing camera of a vehicle to (i) detect other vehicles as bounding boxes (\emph{vehicle detection} via YOLO~\cite{redmon2016you}) and (ii) identify the ego-lane boundaries (\emph{ego-lane detection}). Outputs of these two modules are fed into the third module called \emph{target vehicle selection}, which is as a neural-network based classifier that reports either the index of the bounding box where the target vehicle is located, or a special class for ``no target vehicle''.

As the target vehicle selection neural network takes a fixed number of bounding boxes, one undesired situation of the neural network appears when the network outputs the existence of a target vehicle with index~$i$,  but the corresponding $i$-th input does not contain a vehicle bounding box (marked with ``E" in Figure~\ref{fig:acc}). For the snapshot in Figure~\ref{fig:acc}, the neural network should not output box~$9$ or~$10$ as there are only~$8$ vehicle bounding boxes.
The below code snippet shows how to verify if the unwanted behavior happens in a neural network (\texttt{net}). The undesired property is encoded into two sets of linear constraints, the \texttt{inputConstraints} denotes box~$9$ and~$10$ do not exist and the \texttt{riskProperty} denotes the neural network selects box~$9$ or~$10$ as the final result. The \textit{verify()} function verifies whether if there exists an input that is contained in the specified minimum and maximum bound (\texttt{inputMinBound}, \texttt{inputMaxBound}) while satisfying \texttt{inputConstraints}, but feeding the input to the neural network 
produces an output that satisfies \texttt{riskProperty}.

 \vspace{2mm}
\fbox{\begin{minipage}{23em}
\begin{small}
\texttt{import staticanalysis as sa \\
... \\
sa.verify(inputMinBound, inputMaxBound, net,           inputConstraints, riskProperty)
}
\end{small}
\end{minipage}}

\vspace{2mm}

\paragraph{Perturbation Loss over German Traffic Sign Recognition Network} The last  example is to analyze a neural network trained under the German Traffic Sign Recognition Benchmark~\cite{stallkamp2011german} with the goal of classifying various traffic signs. With \textsf{nn-dependability-kit}, one can apply the \emph{perturbation loss metric} using the below code snippet, in order to understand the robustness of the network subject to known perturbations. 
 
 \vspace{2mm}
\fbox{\begin{minipage}{23em}
\begin{small}
\texttt{import PerturbationLoss as pbl\\
m = pbl.Perturbation\_Loss\_Metric() \\
... \\
m.addInputs(net, image, label) \\
... \\
m.printMetricQuantity("AVERAGE\_LOSS") 
}
\end{small}
\end{minipage}}
 \vspace{2mm}

As shown in the center of Figure~\ref{fig:perturbation}-a, the original image of ``end of no overtaking zone" is perturbed by \textsf{nn-dependability-kit} using seven methods. The application of Gaussian noise changes the result of classification, where the confidence of being ``end of no overtaking zone"  has dropped~$83.4\%$ (from the originally identified~$100\%$ to~$16.6\%$). When applying perturbation systematically over the complete data set, the result is summarized in Figure~\ref{fig:perturbation}-b, where one concludes that the network is robust against haze and fog (the maximum probability reduction is close to~$0\%$) but should be improved against FGSM attacks~\cite{szegedy2013intriguing} and snow.  
  
\section{Outlook}

The journey of \textsf{nn-dependability-kit} originates from an urgent need in creating a rigorous approach for engineering learning-enabled autonomous driving systems. We conclude this paper by highlighting some of our ongoing activities:
(i) Transfer the developed safety concept to standardization communities. (ii) Investigate how assume-guarantee reasoning techniques can make  formal verification scalable. (iii) Extend recent results in safe perturbation bounds~\cite{tsuzuku2018lipschitz,weng2018towards,DBLP:journals/corr/abs-1805-12514,DBLP:journals/corr/abs-1810-07481} to support decisions beyond classification.

\section*{Acknowledgment} As of June 2019, the work was supported by DENSO-fortiss research project \emph{Dependable AI for automotive systems},  Audi-fortiss research project \emph{Audi Verifiable AI}, and Chinese Academy of Science under grant no. 2017TW2GA0003.



%

\end{document}